\documentclass[conference]{IEEEtran}
\IEEEoverridecommandlockouts
\usepackage{cite}
\usepackage{amsmath,amssymb,amsfonts}
\usepackage{algorithmic}
\usepackage{graphicx}
\usepackage{textcomp}
\usepackage{xcolor}
\usepackage{hyperref}
\usepackage{booktabs}
\usepackage{colortbl}
\usepackage{multirow}
\usepackage{siunitx}
\usepackage{relsize}
\def\BibTeX{{\rm B\kern-.05em{\sc i\kern-.025em b}\kern-.08em
    T\kern-.1667em\lower.7ex\hbox{E}\kern-.125emX}}
    
\def\authorname#1{{\em #1}\\} 
\def\address#1{{#1}\\}       

\begin{document}

\title{TAPO: Task-Referenced Adaptation for Prompt Optimization
}




\author{
    \authorname{Wenxin Luo\textsuperscript{*}, 
    Weirui Wang\textsuperscript{*}, 
    Xiaopeng Li\textsuperscript{*}, 
    Weibo Zhou, 
    Pengyue Jia, 
    Xiangyu Zhao\textsuperscript{\textdagger}}
    \address{City University of Hong Kong}
    \{wenxinluo5-c, wrwang8-c, xiaopli2-c\}@my.cityu.edu.hk, xianzhao@cityu.edu.hk
    \thanks{\textsuperscript{*}Equal contribution. \quad \textsuperscript{\textdagger}Corresponding author.}
}

\maketitle

\begin{abstract}
Prompt engineering can significantly improve the performance of large language models (LLMs), with automated prompt optimization (APO) gaining significant attention due to the time-consuming and laborious nature of manual prompt design. However, much of the existing work in APO overlooks task-specific characteristics, resulting in prompts that lack domain specificity and are not well-suited for task-specific optimization. In this paper, we introduce TAPO, a multitask-aware prompt optimization framework composed of three key modules. First, a task-aware metric selection module is proposed to enhance task-specific prompt generation capabilities. Second, we present a multi-metrics evaluation module to jointly evaluate prompts from multiple perspectives. Third, an evolution-based optimization framework is introduced for automatic prompt refinement, which improves adaptability across various tasks. Extensive experiments on six datasets demonstrate the effectiveness of our approach, and our code is publicly available\footnote{\href{https://github.com/Applied-Machine-Learning-Lab/TAPO}{https://github.com/Applied-Machine-Learning-Lab/TAPO}}.
\end{abstract}


\begin{figure*}[!h]
	\centering
	\includegraphics[width=\linewidth]{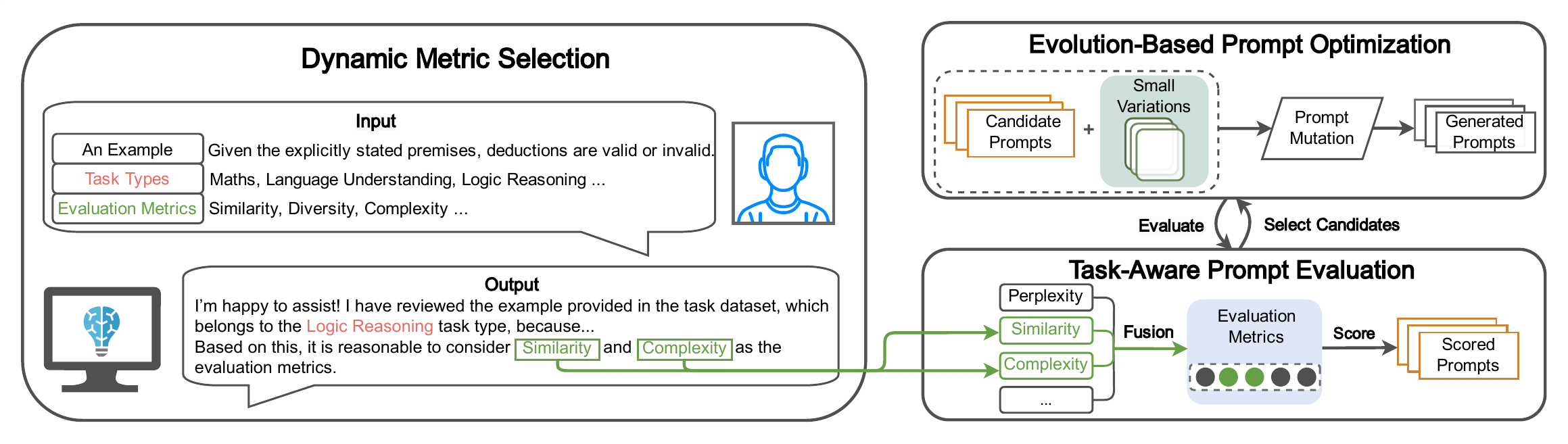}
	\caption{The framework of TAPO. For {\bf Dynamic Metric Selection}, We provide a task dataset example for the LLM to select metrics and assign weights based on priority, creating task-specific evaluation metrics for {\bf Task-Aware Prompt Evaluation}. We employ a tournament selection algorithm for {\bf Evolution-Based Prompt Optimization} to select and mutate the better-performing prompts, adding task-adapted prompts to the candidates.}
	\label{fig:workflow}
\end{figure*}

\section{Introduction}

Prompt engineering plays a critical role in improving the performance of large language models (LLMs)\cite{zhao2023survey}. However, manually constructing prompts is both time-consuming and labor-intensive. Thus, automated prompt optimization~\cite{zhou2022large} has been introduced as a more systematic and efficient approach. Among current approaches, models such as TEMPERA~\cite{Zhang2022TEMPERA} leverage reinforcement learning to dynamically adapt and optimize prompts. Bayesian optimization techniques offer a probabilistic framework for prompt refinement, while in-context learning integrates examples directly into the prompts, as seen in models such as Voke-k~\cite{su2022selective} and Auto-CoT~\cite{zhang2022automatic}. These methods illustrate the growing trend toward more sophisticated prompt engineering strategies.

However, these methods face two primary limitations. First, current prompt evaluation techniques predominantly rely on a single metric, which hinders a comprehensive assessment. For example, PromptBreeder\cite{fernando2023promptbreeder} and APE\cite{zhou2022large} employ a single similarity metric for fitness measurement, limiting their ability to logically improve tasks related to planning. Second, the lack of diverse metrics reduces their versatility, limiting their adaptability to multi-task optimization. For example, GATE~\cite{li2023eliciting} optimizes prompts for role-playing tasks, but exhibits limited scalability to a broader range of tasks. Similarly, while certain prompt strategies focused on machine translation~\cite{noo2024translate} can improve language proficiency, their effectiveness in other essential language tasks, such as communication and reasoning, is diminished.

To address the limitations listed above, we propose Task-Referenced Adaptation for Prompt Optimization (TAPO), a task-aware framework that dynamically selects task-related metrics and automates a task-adaptive prompt evaluation and generation process to facilitate prompt evolution. The framework comprises three key components. The module {\bf Dynamic Metric Selection} enables the LLM to choose relevant metrics according to different tasks and assigns weights based on their priority, establishing task-adapted evaluation metrics for the subsequent stage {\bf Task-Aware Prompt Evaluation}. In the {\bf Evolution-Based Prompt Optimization} module, we use a systematic selection mechanism to iteratively select and mutate high-performing prompts, continuously refining them for improved task-specific performance. In summary, the major contributions of this paper are listed as follows.

\begin{itemize}
    \item We propose Task-Referenced Adaptation for Prompt Optimization (TAPO), an innovative approach that dynamically generates task-specific strategies to enhance multi-task performance and promote generalization across diverse tasks.
    \item A novel task-aware metrics selection and a prompt evaluation module are developed to guide LLMs in generating results that better align with task requirements.
    \item Extensive experiments conducted on six public datasets validate the significance of TAPO's model components and its versatility across diverse tasks.
\end{itemize}

\section{Methodology}


In this section, we introduce the TAPO framework, a dynamic self-improvement method that enhances task-specific performance by optimizing prompts through the selection and weighting of evaluation metrics based on the unique characteristics of each task.

\subsection{Framework Overview}

TAPO's core innovation lies in its multi-objective optimization, which balances criteria like accuracy, fluency, and diversity. As shown in Figure~\ref{fig:workflow}, TAPO integrates LLMs into key components, including task identification, metric selection, and prompt optimization. Dynamically adapts to various tasks by selecting appropriate evaluation metrics and iteratively refining prompts through an adaptive feedback loop, thus improving task-specific performance.

The process begins with task classification, where the LLM identifies the type of task. TAPO then selects relevant metrics, such as similarity and complexity, to guide prompt design and evaluation. High-performing prompts are iteratively refined through mutation and selection, ensuring continuous improvement. This adaptive process makes TAPO flexible and effective in diverse tasks.

\subsection{Dynamic Metric Selection}

Different tasks require different evaluation criteria, and fixed metrics often fail to capture nuanced demands such as precision, creativity, or logical consistency. TAPO optimizes prompts by dynamically selecting and weighting task-specific evaluation metrics. The process begins with task classification, where the LLM-driven module identifies the task type (e.g., reasoning, language, real-world problem) and selects relevant metrics. For factual tasks, similarity ensures accuracy, while creative tasks emphasize diversity to avoid repetition. Metrics such as complexity assess fluency, while both perplexity and logical consistency are crucial for advanced reasoning, dialogue, and decision-support systems. This approach enables TAPO to adapt to various tasks, ensuring optimal performance across multiple dimensions.

\subsection{Task-Aware Prompt Evaluation}

To evaluate and adapt prompts for various tasks, we propose a prompt evaluation module with two components: metric fusion and dynamic weight adjustment. After selecting the evaluation metrics, TAPO combines them into a final scoring function to comprehensively assess the performance of the task. The scoring function is defined as:
\begin{equation}
    S(\mathcal{P}) = \sum_{i=1}^{n} w_i \cdot M_i(\mathcal{P})
\end{equation}
where \(\mathcal{P}\) is the optimized prompt, \(w_i\) is the weight of the \(i\)-th metric, \(M_i(\mathcal{P})\) denotes the score for the \(i\)-th metric, and \(S(\mathcal{P})\) represents the overall score for \(n\) metrics.

TAPO integrates similarity, diversity, perplexity, and complexity as metrics to balance accuracy, creativity, and fluency. Adjusts the weight of each metric based on task requirements, prioritizing similarity for precision tasks and enhancing diversity and perplexity for creative tasks.

\subsection{Evolution-Based Prompt Optimization}

Traditional prompt optimization methods often stagnate in local optima, limiting their ability to explore better alternatives. TAPO addresses this limitation by refining prompts through evolutionary strategies, leveraging mutation, and selection for continuous improvement. 

During initialization, TAPO generates prompts by integrating random thinking styles with the problem description, which are subsequently processed by the LLM. In self-evolution processes, small variations, such as "breaking the task into steps", are selected from a predefined strategy library to combine with candidate prompts. These combinations are then processed through mutation operators to generate evolved prompts. During each iteration, the performance evaluation is conducted using the multi-metric scoring function mentioned above. TAPO applies tournament selection to filter candidates, ensuring improved task-specific results. This iterative process continues for multiple cycles, dynamically refining the prompts until they achieve the desired performance or reach a predefined iteration limit, ensuring continuous optimization.


\begin{table*}[htbp]
\centering
\scriptsize
\caption{\larger[1] Performance comparison on different datasets with GPT-3.5-turbo and GPT-4o (Similarity Scores).}
\resizebox{\textwidth}{!}{
\begin{tabular}{l S[table-format=2.2] S[table-format=2.2] S[table-format=2.2] S[table-format=2.2] S[table-format=2.2]
                  S[table-format=2.2] S[table-format=2.2] S[table-format=2.2] S[table-format=2.2] S[table-format=2.2]}
\toprule
\multirow{2}{*}{\textbf{Dataset}} & \multicolumn{5}{c}{\textbf{GPT-3.5-turbo}} & \multicolumn{5}{c}{\textbf{GPT-4o}} \\
\cmidrule(lr){2-6} \cmidrule(lr){7-11}
 & {COT} & {APE} & {PE2} & {PB} & {TAPO} & {COT} & {APE} & {PE2} & {PB} & {TAPO} \\
\midrule
BBH(23 tasks) & 66.68 & 68.10 & 63.57 & \underline{68.17} & \textbf{69.28}* & 74.18 & 74.83 & 77.05 & \underline{79.90} & \textbf{80.51}* \\
GSM8K & \textbf{83.70}* & 81.99 & 78.63 & 82.45 & \underline{83.40} & 85.49 & 79.79 & 83.37 & \textbf{88.61}* & \underline{88.40} \\
AddSub & 58.61 & 57.04 & 68.10 & \underline{82.11} & \textbf{88.15}* & \textbf{100.00}* & \underline{97.92} & 95.78 & 97.62 & 96.32 \\
MultiArith & 69.00 & \underline{86.36} & 83.56 & 85.26 & \textbf{89.26}* & 100.00 & 100.00 & \underline{97.92} & 100.00 & \textbf{100.00} \\
SingleEQ & 61.91 & 63.14 & 78.92 & \underline{82.11} & \textbf{89.06}* & 96.12 & \underline{96.78} & 89.63 & 96.74 & \textbf{97.83}* \\
SVAMP & \textbf{94.38}* & \underline{93.10} & 91.02 & 90.44 & 92.72 & 93.23 & 95.71 & 94.81 & \underline{99.72} & \textbf{100.00}* \\
\bottomrule
\\[-1.5ex]
\multicolumn{11}{l}{``*'' indicates significance level test $p<0.05$,  the suboptimal results are \underline{underlined}.}
\end{tabular}}
\label{table:result1}

\end{table*}

\section{Experiment}

In this section, we present experimental settings and outline the design of our experiments to address the following research questions: \textbf{RQ1:} How does our model perform compared to state-of-the-art approaches? \textbf{RQ2:} How effectively does our model adapt to different types of tasks? \textbf{RQ3:} Does our framework maintain consistent performance across open-source LLMs? \textbf{RQ4:} What is the impact of individual components on overall performance?

\subsection{Experiment Settings}

\textbf{Datasets.} To evaluate our method, we use a range of datasets focused on mathematical reasoning and multi-task problem solving, including AddSub \cite{hosseini2014learning}, MultiArith \cite{roy2016solving}, and SingleEQ \cite{koncel2015parsing} for arithmetic reasoning, as well as SVAMP \cite{patel2021nlp} and GSM8K \cite{cobbe2021training} for multi-step problem solving. Furthermore, we incorporate BIG-Bench Hard (BBH) \cite{suzgun2022challenging}, a dataset comprising 23 diverse and challenging tasks, including logical reasoning and common sense understanding, to ensure a comprehensive evaluation.

\textbf{Baselines.} We compare TAPO against the following baseline methods: (a) Zero-Shot CoT \cite{wei2022chain}, which generates reasoning steps in a zero-shot manner; (b) APE \cite{zhou2022large}, a method that initializes multiple prompt candidates from a base prompt and selects the best one based on development set performance; (c) PE2 \cite{ye2023prompt}, a two-step prompting approach that iteratively generates and refines candidate prompts through evaluation; and (d) PromptBreeder (PB)\cite{fernando2023promptbreeder}, a self-referential optimization framework that refines prompts by leveraging redescriptions to improve downstream task performance.

\textbf{Experiment Details.}  
In our experiments, we utilize the following language models: \texttt{GPT-3.5-turbo-0125} \cite{openai_gpt35turbo}, \texttt{GPT-4o-2024-08-06}\cite{hurst2024gpt}, and \texttt{Llama3-8B-Instruct}\cite{dubey2024llama}. The first two models are accessed through the OpenAI API, while \texttt{Llama3} is deployed using the NVIDIA API. To ensure consistency between tasks, the temperature is set to 0.1. The evaluation of generated text employs metrics for similarity, fluency, diversity, and complexity. Similarity is assessed using cosine similarity, calculated via the \texttt{all-MiniLM-L6-v2}\cite{reimers2019sentence} model, to measure the semantic alignment between the generated and reference texts. Fluency is evaluated through perplexity scores derived from the \texttt{gpt2-large}\cite{radford2019language} model, where lower values indicate more coherent and grammatically accurate outputs. Diversity is used to quantify lexical variety by calculating the proportion of unique n-grams, with higher scores reflecting reduced repetition. Complexity is assessed by analyzing text length, syntactic structures, and logical reasoning steps.

\subsection{Overall Performance (RQ1)}
TAPO consistently outperforms baseline methods by dynamically selecting and weighting task-specific metrics. In arithmetic reasoning tasks such as AddSub and MultiArith, TAPO achieves similarity scores of 88.15\% and 89.26\% in \texttt{GPT-3.5-turbo}, respectively, demonstrating a clear advantage over static methods like CoT\cite{wei2022chain} and APE\cite{zhou2022large}. For multi-step reasoning tasks such as GSM8K, TAPO reaches 88.40\% on \texttt{GPT-4o}, just shy of the best score at 88.61\%. In BBH, TAPO achieves 80.51\% in \texttt{GPT-4o}, slightly exceeding the next best method at 79.90\%.

Although TAPO does not always achieve the highest score, such as in SVAMP where it reaches 92.72\% on \texttt{GPT-3.5-turbo} compared to 94.38\%, it consistently ranks among the best methods, showing strong adaptability in various tasks. These results, summarized in Table~\ref{table:result1}, underscore TAPO’s effectiveness in optimizing task-specific prompts for a wide range of language and reasoning tasks.

\subsection{Task-Specific Prompt Performance (RQ2)}
\vspace*{-2mm}
\begin{table}[htbp]
\caption{\normalsize Effect of Different Prompt Optimization Methods.}
\scriptsize
\centering
\small
\begin{tabular}{>{\centering\arraybackslash}m{2cm} m{6cm}}
\toprule
\textbf{Method} & \textbf{Math Reasoning Prompt} \\ 
\midrule
Zero-shot CoT & Let’s think step by step. \\ \hline
APE & Solve arithmetic word problems. \\ \hline
PB & Subtract 2 from 8 to find how many kittens Joan has now. \newline Answer: 8 - 2 = 6 \newline Advice: Correct subtraction. Well done! \\ \hline
\rowcolor{green!20} TAPO & Break the problem into smaller parts. Identify key elements, use diagrams or rephrase, and remove unnecessary information. \\ 
\end{tabular}

\begin{tabular}{>{\centering\arraybackslash}m{2cm} m{6cm}}
\toprule
\textbf{Method} & \textbf{Translation Error Detection Prompt} \\ 
\midrule
Zero-shot CoT & Let’s think step by step. \\ \hline
APE & Identify translation errors based on specific categories like Named Entities, Numerical Values, and Modifiers. \\ \hline
PB & Become a fearless error detective with a magnifying glass, finding quirky translation missteps in a whimsical German fairytale. \\ \hline
\rowcolor{green!20} TAPO & Improve translation error detection by creating a structured framework to categorize errors, focusing on common issues, and using iterative testing with feedback to refine strategies. \\ 
\toprule
\end{tabular}

\label{table:task_aware_performance}
\end{table}

TAPO's ability to tailor prompts for specific tasks significantly enhances performance across different domains, as illustrated in Table~\ref{table:task_aware_performance}. For math reasoning tasks like AddSub, TAPO emphasizes reasoning and computational steps, offering a customized approach that outperforms general methods like zero-shot CoT, which uses the generic prompt \texttt{"Let's think step by step"}.  For translation tasks, TAPO excels by using a systematic framework for error categorization and iterative feedback, in contrast to the less structured methods of APE or PromptBreeder. This task-specific design ensures that TAPO consistently delivers superior results by adapting the prompts to the unique requirements of each task, improving both clarity and performance in mathematical and translation error detection tasks.

\subsection{Open-Source LLM Performance (RQ3)}

\begin{figure}[htbp]
\centerline{\includegraphics[width=\linewidth]{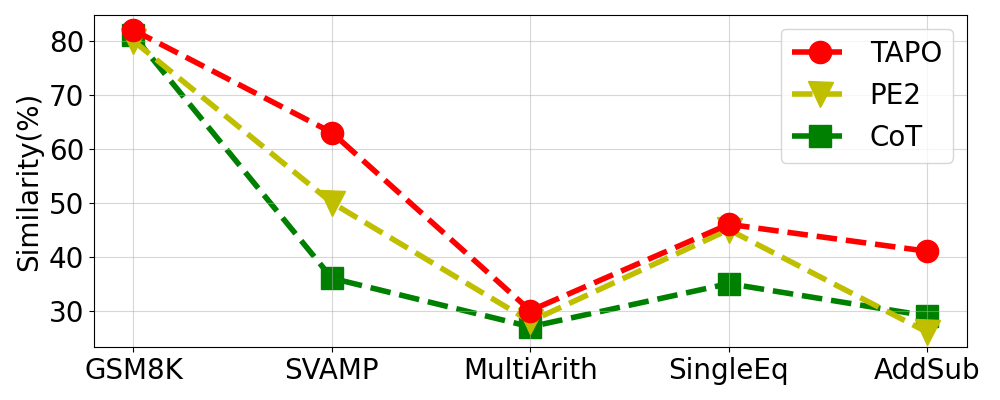}}
\vspace{-2mm}
\caption{Performance Comparison with Llama3-8B-Instruct.}
\label{fig:similarity_scores}
\end{figure}

When evaluating open-source large language models, such as \texttt{Llama3-8B-Instruct}, TAPO consistently outperforms baseline methods, including CoT and PE2, even on tasks that require precise formatting and multi-step reasoning. As illustrated in Figure~\ref{fig:similarity_scores}, while \texttt{Llama3-8B-Instruct} struggles to maintain the correct output formats on datasets such as MultiArith and AddSub, TAPO enables the model to achieve significantly better results. In tasks like GSM8K, which emphasize multi-step reasoning, TAPO further improves the LLM’s performance, narrowing the gap with state-of-the-art models. On average, TAPO improved similarity scores in math reasoning tasks by 10.2\% compared to CoT and 6.2\% compared to PE2. These results demonstrate that TAPO's optimization enhances performance even in open-source LLMs, making it effective across different model architectures.

\subsection{Ablation Study (RQ4)}

\begin{table}[htbp]
\caption{Ablation study of TAPO with gpt-3.5-turbo\\(Similarity Scores).}
\begin{center}
\resizebox{\columnwidth}{!}{
\begin{tabular}{lccccc}
\toprule
\textbf{Method} & \textbf{AddSub} & \textbf{SingleEQ} & \textbf{SVAMP} & \textbf{MultiArith} & \textbf{GSM8K} \\
\midrule
\textbf{TAPO (Full)} & \textbf{88.15\%} & \textbf{89.06\%} & \textbf{92.72\%} & \textbf{89.26\%} & \textbf{83.40\%} \\

\textbf{w/o PO} & \underline{87.24\%} & \underline{82.75\%} & \underline{85.38\%} & \underline{83.04\%} & 81.82\% \\

\textbf{w/o MS} & 82.60\% & 75.91\% & 80.14\% & 78.58\% & \underline{82.41\%} \\
\bottomrule

\end{tabular}}
\label{tab:ablation_gpt35}
\end{center}
\end{table}

We conducted an ablation study to evaluate the key components of TAPO, as shown in Table~\ref{tab:ablation_gpt35}, focusing on removing prompt optimization (PO) and multi-metric scoring (MS). In the \textbf{w/o PO} variant, we replaced the optimization of task-specific prompts with a generic approach, which resulted in a performance decline across all datasets, highlighting the importance of refined prompt generation of TAPO. Similarly, in the \textbf{w/o MS} variant, using a single-metric method instead of the multi-metric approach led to a significant performance drop, particularly in datasets such as SigleEQ and MultiArith, highlighting the crucial role of multi-metric evaluation in improving task-specific results.

\section{Related Work}

The optimization of prompts has become central to the application of LLMs in areas such as agents \cite{cai2024agentir,li2023agent4ranking}, RAG \cite{jia2024bridging,jia2024g3}, IR \cite{li2024syneg,jia2023mill}, RecSys~\cite{liu2024moe, liu2024llm, li2023hamur, gao2024hierrec, jia2024d3, li2024scenario, liu2024large1, zhang2024llm}, etc., and it has recently gained significant attention due to the time-consuming and labor-intensive process of manual prompt tuning. Several approaches have been proposed, including leveraging feedback from LLMs and iterative improvements~\cite{zhou2022large}, reinforcement learning-based methods~\cite{deng2022rlprompt}, Bayesian optimization techniques~\cite{liu2024large}, in-context learning~\cite{ye2023prompt}, and evolutionary algorithms~\cite{fernando2023promptbreeder}, among others. However, current prompt evaluation methods rely on uniform metrics that lack adaptability and fail to optimize for different task-specific objectives. In this paper, we propose a task-aware automatic prompt optimization framework that dynamically selects evaluation metrics based on task-specific feedback, thereby improving the generalizability of prompt optimization across multiple tasks.


\section{Conclusion}

We propose a novel framework, TAPO, to improve prompt generation and enhance the adaptability of LLMs to diverse tasks. TAPO leverages a comprehensive set of evaluation metrics, dynamically adjusting them based on task requirements, while an adaptive feedback loop iteratively refines the prompts to ensure continuous improvement. Extensive experiments on various datasets demonstrate that TAPO consistently outperforms existing methods, offering superior performance and versatility across different models and task types, ranging from arithmetic reasoning to multistep problem solving, creative generation, and logical reasoning challenges.

\section*{Acknowledgment}

This research was partially supported by Research Impact Fund (No.R1015-23), Collaborative Research Fund (No.C1043-24GF), APRC - CityU New Research Initiatives (No.9610565, Start-up Grant for New Faculty of CityU), CityU - HKIDS Early Career Research Grant (No.9360163), Hong Kong ITC Innovation and Technology Fund Midstream Research Programme for Universities Project (No.ITS/034/22MS), Hong Kong Environmental and Conservation Fund (No. 88/2022), and SIRG - CityU Strategic Interdisciplinary Research Grant (No.7020046), Huawei (Huawei Innovation Research Program), Tencent (CCF-Tencent Open Fund, Tencent Rhino-Bird Focused Research Program), Ant Group (CCF-Ant Research Fund, Ant Group Research Fund), Alibaba (CCF-Alimama Tech Kangaroo Fund No. 2024002), CCF-BaiChuan-Ebtech Foundation Model Fund, and Kuaishou.

\vfill\pagebreak 

\bibliographystyle{IEEEtran}
\bibliography{refs}

\end{document}